# Decomposition of Water Demand Patterns Using Skewed Gaussian Distributions for Behavioral Insights and Operational Planning

Roy Elkayam[†]

[†]Mekorot Water Company, Lincoln Street, Tel-Aviv – Yafo, 6492105, Israel.

**Abstract**

This study presents a novel approach for decomposing urban water demand patterns using Skewed Gaussian Distributions (SGD) to derive behavioral insights and support operational planning. Hourly demand profiles contain critical information for both long-term infrastructure design and daily operations, influencing network pressures, water quality, energy consumption, and overall reliability. By breaking down each daily demand curve into a baseline component and distinct peak components, the proposed SGD method characterizes each peak with interpretable parameters, including peak amplitude, timing (mean), spread (duration), and skewness (asymmetry), thereby reconstructing the observed pattern and uncovering latent usage dynamics. This detailed peak-level decomposition enables both operational applications, e.g. anomaly and leakage detection, real-time demand management, and strategic analyses, e.g. identifying behavioral shifts, seasonal influences, or policy impacts on consumption patterns. Unlike traditional symmetric Gaussian or purely statistical time-series models, SGDs explicitly capture asymmetric peak shapes such as sharp morning surges followed by gradual declines, improving the fidelity of synthetic pattern generation and enhancing the detection of irregular consumption behavior. The method is demonstrated on several real-world datasets, showing that SGD outperforms symmetric Gaussian models in reconstruction accuracy, reducing root-mean-square error by over 50% on average, while maintaining physical interpretability. The SGD framework can also be used to construct synthetic demand scenarios by designing daily peak profiles with chosen characteristics. All implementation code is publicly available at: https://github.com/Relkayam/water-demand-decomposition-sgd

**Keywords:** Skewed Gaussian Decomposition, Urban Water Demand, Demand Pattern Reconstruction

# 1 Introduction

Water distribution systems must reliably provide safe drinking water amid highly variable demand. Characterizing and modeling water demand patterns are crucial for effective system design, operation, and management (Blokker et al. 2017). Temporal variations in demand influence system pressure, water quality, energy use, and service reliability (Billings and Jones 2008; Herrera et al. 2010). Therefore, accurate mathematical representations of diurnal demand cycles are needed for applications ranging from hydraulic simulation to real-time operational control. However, water demand patterns exhibit complex structures with recurring daily cycles, irregular multi-modal peaks, and stochastic fluctuations that pose challenges for utilities in both short-term operations and long-term planning (Billings and Jones 2008).

While extensive research exists on water demand forecasting (Alvisi et al. 2024; Donkor et al. 2014; Ghalehkhondabi et al. 2017; Pacchin et al. 2019; Romano and Kapelan 2014), relatively few studies have focused on explicitly decomposing observed demand profiles into their constituent peak events. Most prior decomposition approaches have employed either symmetric or linear segmentation techniques – such as time-series segmentation (Moretti et al. 2022; Shvartser et al. 1993), Fourier analysis (Zhou et al. 2002), or machine learning pattern extraction (Antunes et al. 2018), which may overlook critical peak-level details or sacrifice physical interpretability. Additionally, many studies emphasize disaggregated end-use or household-level demand patterns (Cominola et al. 2023; Gargano et al. 2016; Herrera et al. 2010; Di Mauro et al. 2020; Pesantez et al. 2020; Worthington and Hoffman 2008). In contrast, top-down methods for reconstructing aggregated demand curves using a limited set of parameters remain underdeveloped, limiting utilities' ability to interpret demand at scale and synthesize realistic operational scenarios.

The ability to generate synthetic demand patterns is especially valuable when observed data are limited, proprietary, or when exploring extreme and hypothetical scenarios (Niknam et al. 2022). Synthetic profiles allow utilities to stress-test infrastructure designs and operational strategies under varied conditions (Behzadian et al. 2014). This study addresses the above needs by introducing a new decomposition framework based on Skewed Gaussian Distributions. To our knowledge, this is the first application of SGDs to decomposing hourly water demand profiles. Unlike traditional Gaussian models that assume perfectly symmetric (bell-shaped) peaks (McKenna et al. 2014), or standard ARIMA-type time-series models that capture trends and autocorrelation but not peak shape, SGD introduces a skewness parameter that allows each peak to be modeled with asymmetry. This added flexibility enables representation of common behaviors such as a rapid early-morning demand surge followed by a slower decline, or conversely a gradual ramp-up and sharp cutoff – patterns that symmetric models cannot fully capture. This method's novelty lies in capturing both the asymmetry and physical interpretability of demand peaks using a compact, parameterized formulation suitable for operational deployment. The main advantage of using SGDs lies in their ability to preserve key peak characteristics (timing, magnitude, and shape) while explicitly modeling behavioral asymmetry, such as differing rates of rise and fall in usage. This can improve the realism of synthetic demand generation, enhance detection of abnormal consumption patterns, and support more reliable hydraulic simulations under variable or extreme conditions. Moreover, the skewness parameters can reveal latent behavioral signatures within aggregated demand data, providing new insights into user dynamics at the system level.

Skewed Gaussian functions have proven useful in other domains (e.g. environmental modeling and neurophysiology) for representing asymmetric signals (Azzalini 1985; Grün et al. 2010), yet their potential remains largely unexplored in urban water demand modeling, where demand peaks are often asymmetric and information-rich. This paper presents an SGD-based decomposition methodology and applies it to six published hourly demand patterns from various contexts. The research compares the reconstructed demand curves to the original data to assess accuracy and peak-preservation, and contrasts the performance of the skewed Gaussian model with a baseline that uses only symmetric Gaussians. By isolating and modeling individual peak events using SGDs, the aim is to provide a scalable framework that balances physical interpretability with computational robustness, suitable for both research and operational applications.

## 2  Methodology

This section outlines the proposed methodology for decomposing an hourly water demand pattern into a sum of Skewed Gaussian components plus a baseline term. The approach is designed to capture distinct peak characteristics – such as amplitude, timing, duration, and asymmetry – while maintaining interpretability and operational relevance. A complete implementation, including preprocessing, optimization, and plotting scripts, is available in the GitHub repository for reproducibility (Elkayam 2025)

## 2.1 Skewed Gaussian Model

To model the asymmetric peaks commonly observed in water demand, the SGD is employed. A skewed Gaussian extends the standard normal (Gaussian) distribution by introducing a shape parameter $\alpha$ to control asymmetry. The probability density function (adapted as a flexible peak shape) can be defined as:

Eq.1 $$f(x; A, \mu, \sigma, \alpha) = A \cdot exp\left(-\frac{(x-\mu)^2}{2\sigma^2}\right) \cdot \left[1 + erf\left(\alpha \cdot \frac{x-\mu}{\sqrt{2}\sigma}\right)\right]$$

Here $A$ is the peak amplitude (in flow units, e.g. m³/h or gallons/h), $\mu$ is the peak time location (hour of day), $\sigma$ is the peak width (spread, in hours), and $\alpha$ is the dimensionless skewness parameter. The function *erf* is the Gauss error function. When $\alpha = 0$, this reduces to a symmetric Gaussian (normal) curve centered at $\mu$. A positive $\alpha$ skews the peak to the right (a sharp rise followed by a slower decline), whereas a negative $\alpha$ skews it to the left (gradual rise, rapid drop-off). This formulation allows flexible modeling of both symmetric and asymmetric peak shapes commonly seen in real consumption data.

## 2.2 Composite Demand Model

The total demand profile D(t) for a given day is modelled as the combination of a baseline demand and several peak components, each represented by a SGD. Mathematically,

Eq.2 $$D(t) = C_{base} + \sum_{i=1}^{n_{peaks}} f(t; A_i, \mu_i, \sigma_i, \alpha_i)$$

In this formulation, $C_{base}$ is a constant base demand representing the time-independent background flow (e.g. from leakage or continuous industrial usage). The summation term adds up $n$ skewed

Gaussian peak functions corresponding to distinct usage events or peak periods. This composite model is applied at the system level (e.g. the output of a distribution reservoir or a district metering area), rather than at a single household. Each peak $i$ has its own parameters $(A_i, \mu_i, \sigma_i, \alpha_i)$, which will be determined through the decomposition algorithm.

## 2.3 Peak Detection Algorithm

Significant peaks were identified using a first derivative-based algorithm. Hourly water demand data were differentiated to identify points of inflection where the flow transitions from increasing to decreasing. These inflection points were flagged as potential peaks based on the sign change of the first derivative. To ensure coverage of corner cases and detect sustained high-demand periods:

**Endpoint Correction:** The first and last timepoints were handled separately to ensure local maxima at boundaries were properly detected. The first point was flagged as a peak if it was greater than the second, and the last point if it was greater than the penultimate.

**Plateau Detection:** Plateau regions, defined as consecutive hours with no change in flow (derivative = 0), were further evaluated. Plateaus spanning two or more consecutive hours were flagged as peaks, recognizing sustained demand levels that may not present sharp slopes but are operationally relevant.

## 2.4 Optimization Approach

Once peak locations are identified, the Skewed Gaussian parameters are fitted to the observed demand data. This is formulated as a constrained nonlinear optimization problem to minimize the discrepancy between the modeled demand and the observed demand. The objective function to be minimized is a loss that combines the goodness-of-fit (Mean Squared Error, MSE) with regularization terms to encourage realistic peak shapes. In particular, the loss function L is defined as:

Eq.3 $$L = \frac{1}{N}\sum_{i=1}^{N}(y_i - \hat{y}_i)^2 + \sum_{j=1}^{n_{peaks}}(\sigma_j - R_1)^2 + R_2 \sum_{j=1}^{n_{peaks}} \alpha_i^2$$

Here $y_i$ is the observed demand at hour $i$ and $\hat{y}_i$ is the model-predicted demand (from Eq. 2) at that hour, and $N=24$ for a daily profile. The second term imposes a penalty on peak widths $\sigma_j$ deviating from an expected value, and the third term penalizes large skewness values $\alpha_j$. Regularization is applied to discourage the formation of unrealistically wide or highly skewed peaks. To this end, a nominal peak width of $R_1 = 2$ hours is used, and the regularization weight for skewness is set to $R_2 = 0.01$, ensuring that extreme parameter values are constrained. The optimization was solved using the L-BFGS-B algorithm (Limited-memory Broyden–Fletcher–Goldfarb–Shanno with simple bounds) (Saputro and Widyaningsih 2017), as implemented in the SciPy library (Virtanen et al. 2020). This quasi-Newton gradient-based method handles bound constraints efficiently and is well-suited for moderate-dimensional problems (Byrd et al. 1995). Each peak contributes four parameters ($A, \mu, \sigma, \alpha$) plus one $C_{base}$, resulting in a total parameter count of $4n + 1$. Sensible parameter bounds were enforced to avoid unrealistic solutions, based on expected operational ranges in typical systems:

- Baseline $C_{base}$: constrained between 0 and the 10th-percentile of the hourly demand (to approximate the non-peak minimum night flow).
- Peak amplitude $A$: constrained to [0, 1.2 × max observed hourly flow]
- Peak width $\sigma$: constrained to: [0.1, 10] hours (allowing for very narrow to very broad peaks).
- Skewness $\alpha$: constrained to [−5, +5] (unitless), to accommodate highly skewed shapes while avoiding numerical instability.

To improve the robustness of the optimization and avoid local minima, a multi-start strategy was adopted. The solver was run multiple times from different initial guesses for the key non-linear parameters (peak width $\sigma$ and skewness $\alpha$), while other parameters were initialized based on rough estimates (e.g. initial $A$ from peak height, initial $\mu$ from detected peak time). Specifically, $\sigma$ was initialized with values in $\{1.0, 2.0, 3.0\}$ hours and $\alpha$ in $\{-1.0, 0.0, +1.0\}$, covering a range of symmetric and mildly skewed shapes. All combinations of these initial values were tested for each peak, and the solution with the lowest objective function value was selected. This multi-start approach typically provided a 12–18% reduction in RMSE (on the training fit) compared to a single-start run, indicating that a better global optimum was found (Gao et al. 2020; György and Kocsis 2011).

## 2.5 Model validation

The fitted models were evaluated using multiple error metrics to quantify reconstruction accuracy on the same data used for fitting, as the objective was descriptive fidelity rather than generalization to unseen days. Root Mean Square Error (RMSE), Mean Absolute Error (MAE), and Maximum Absolute Error were computed between the modeled and observed hourly demands. These metrics were calculated in original flow units for each dataset (i.e., without normalization), to reflect real operational deviations. Additionally, the coefficient of determination $R^2$ was computed to indicate goodness-of-fit. Because each daily pattern was modeled in full - rather than dividing the data into training and testing subsets - these validation metrics reflect how well the decomposition captures the specific shape of the observed daily demand. This supports operational interpretability over predictive performance, as the goal is to extract actionable structure from known patterns rather than forecast future demand. This is consistent with the principles discussed in Ghalehkhondabi et al. (2017) and Donkor et al. (2014), who emphasize the value of descriptive models in water demand analysis when the objective is explanation rather than forecasting. Although no separate train/test split was employed, the reported error metrics still offer a rigorous evaluation of the model's ability to replicate known patterns.

## 2.6 Evaluation Datasets

To test the methodology, it was applied to six distinct real-world datasets (Table 1), each representing different urban contexts and demand behaviors: i) **P1a & P1b**: Two demand profiles from Northern California before and after a drought mandate (Nemati et al. 2023) ii) **P2a & P2b**: Two patterns from East England during the COVID-19 lockdown (Abu-Bakar et al. 2021). And iii) **P3a & P3b**: Two high-volume profiles (daily average ~2440 m³/day) from a European settlement (Plantak et al. 2022). These datasets were selected to represent small-to-large systems, under both routine and disrupted behavioral conditions.

*Table 1: Characteristics of Evaluation Demand Patterns, each pattern corresponds to one day of data:*

| Pattern ID | Units | Flow Range | Mean flow | Source |
|---|---|---|---|---|
| P1a | Gallons/h | (11.7, 23.2) | 13.33 | (Nemati et al. 2023) |
| P1b | Gallons/h | (9.0, 14.8) | 9.3 | (Nemati et al., 2023) |
| P2a | m³/h | (4.0, 108.0) | 47.75 | (Abu-Bakar et al. 2021) |
| P2b | m³/h | (2.0, 34.0) | 20.75 | (Abu-Bakar et al., 2021) |
| P3a | m³/h | (24.4, 170.8) | 101.68 | (Plantak et al. 2022) |
| P3b | m³/h | (12.2, 256.2) | 100.25 | (Plantak et al., 2022) |

## 3 Results

The results of applying the Skewed Gaussian Decomposition (SGD) to each of the six demand patterns are first presented, followed by a comparison with the symmetric Gaussian baseline.

### 3.1 Decomposition Analysis (SGD Model)

The SGD-based algorithm successfully decomposed all six daily demand profiles (P1a, P1b, P2a, P2b, P3a, P3b) into a sum of skewed Gaussian peaks plus a baseline. Figure 1 illustrates the fitted model for each pattern, showing how individual skewed Gaussian components (dashed curves) combine to closely match the observed demand curve (solid blue). The model captures both the major peaks and the minor fluctuations (often below ~10% of the daily maximum amplitude) in each case, with the red line (total modeled demand) closely overlapping the actual data points.

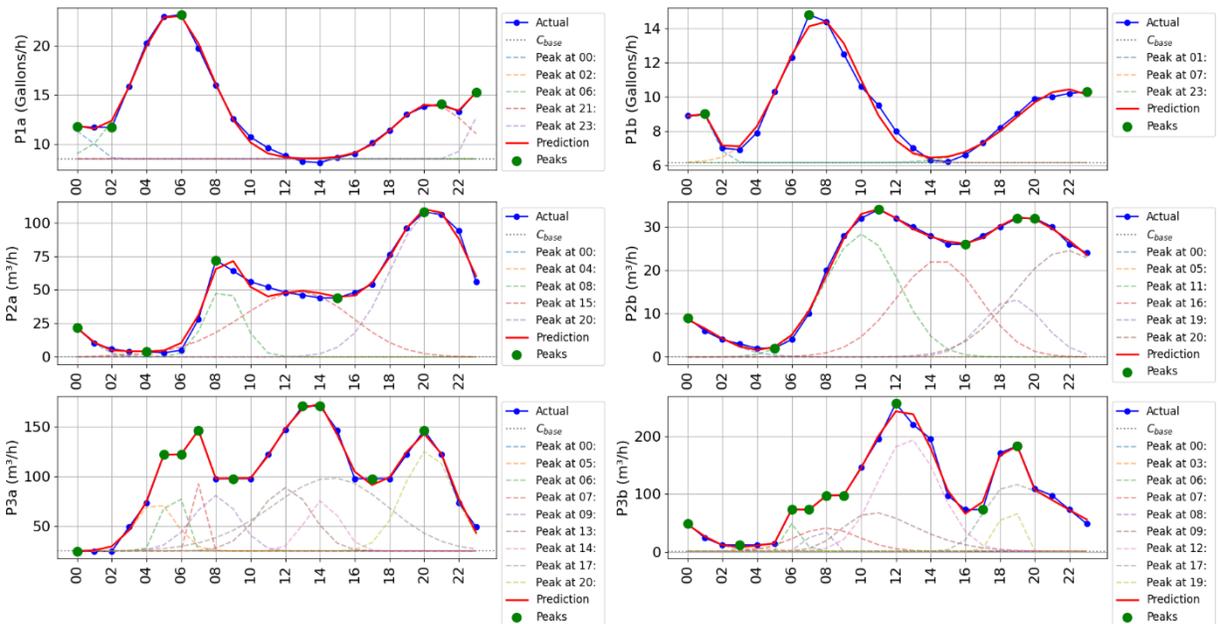

*Figure 1. Observed vs. modeled demand patterns using the SGD decomposition. Each subplot corresponds to one daily pattern (P1a, P1b, P2a, P2b, P3a, P3b). Blue markers show the observed*

*demand. The red line shows the total reconstructed demand from the SGD model. Dashed colored curves represent the individual skewed Gaussian peak components, with a green dot indicating each peak's apex. The baseline component $C_{base}$ is shown as a dotted horizontal line. Time of day (00:00–23:00) is on the x-axis; flow units (gallons/hour or m³/hour) are on the y-axis as indicated for each plot.*

Across these diverse profiles, the algorithm identified between 3 and 9 peaks per day and adjusted their shapes to reflect underlying usage behavior. Table 2 summarizes the model parameters for each pattern, including the number of peaks found ($n$), the fitted baseline $C_{base}$, and the ranges of peak amplitudes, widths, and skewness values. These parameters highlight key differences between patterns. For example, the California patterns (P1a, P1b) each contain 3–5 peaks with moderate amplitudes (up to ~15 units) and relatively narrow widths (most $\sigma < 3$ h), reflecting typical residential diurnal usage. The English lockdown patterns (P2a, P2b) required 5–6 peaks, including sharp spikes (e.g. P2a amplitude ~108 m³/h) and notably asymmetric shapes, skewness $\alpha$ down to –2.0 in P2a, indicating a slow decay after a sharp rise.

The large European system patterns (P3a, P3b) featured the most peaks (9 each) and the widest behavioral range, P3b, for example, included both highly negative and highly positive skewness values (from –3.06 to +4.71), indicating variation in peak shape within a single day. P3a and P3b also exhibited non-zero baselines $C_{base}$ values between approximately 1.5 and 25 m³/h, consistent with continuous night flow typical in larger systems. Even subtle shoulders and secondary bumps in the demand curves were captured by smaller SGD components, underscoring the method's sensitivity to overlapping usage signals.

*Table 2. Skewed Gaussian Model Parameters, Summary of model parameters derived from the skewed Gaussian decomposition for each water demand pattern. Includes number of peaks, base demand level $C_{base}$ [units: gallons/hour or m³/hour depending on pattern], and range (min–max) of peak amplitude [same units], width σ [hours], and skewness α [unitless].*

| Pattern ID | N peaks | $C_{base}$ | Amplitude Range | Sigma Range | Alpha Range |
|---|---|---|---|---|---|
| P1a | 5 | 8.48 | (0.0, 14.57) | (0.73, 2.18) | (-0.37, 0.64) |
| P1b | 3 | 6.16 | (2.74, 7.97) | (1.02, 3.18) | (-0.85, 0.39) |
| P2a | 5 | 0 | (1.7, 107.64) | (1.15, 4.18) | (-2.04, 0.62) |
| P2b | 6 | 0 | (0.0, 25.67) | (1.75, 3.71) | (-1.01, 0.94) |
| P3a | 9 | 24.68 | (0.0, 100.2) | (0.31, 4.35) | (-3.16, 0.2) |
| P3b | 9 | 1.48 | (3.61, 179.96) | (0.46, 4.57) | (-3.06, 4.71) |

The goodness-of-fit of the SGD models was very high in all cases. Table 3 reports the error metrics for each pattern. For four of the six patterns, R² exceeded 0.99 (indicating that over 99% of the variance in demand was explained by the model). Even for the most complex case (P3b), R² was ~0.989 with RMSE on the order of 7.17 units (7.16% of mean demand). Simpler profiles like P1a, P1b, and P2b achieved RMSE under 1 unit (only ~2–3% of mean flow) and near-perfect R². The largest errors tended to occur at the sharpest peaks (e.g. P2a and P3b have the highest max error, around 18 m³/h in P3b which occurred at a single extreme hour). Overall, these results demonstrate that the SGD decomposition can reconstruct the observed demand patterns with high fidelity, successfully capturing both major peak magnitudes and the overall demand volume.

Table 3. Performance Metrics – Skewed Gaussian Model, Model performance metrics for the skewed Gaussian fit for each pattern. RMSE, MAE, and maximum error are in the same units as the pattern (gallons/hour or m³/hour); $R^2$ is unitless.

| Pattern ID | RMSE | RMSE as % of mean | MAE | MAE as % of mean | Max Error | $R^2$ |
|---|---|---|---|---|---|---|
| P1a | 0.28 | 2.14 | 0.22 | 1.61 | 0.68 | 0.996 |
| P1b | 0.32 | 3.41 | 0.25 | 2.73 | 0.69 | 0.982 |
| P2a | 3.54 | 7.4 | 2.68 | 5.61 | 7.15 | 0.988 |
| P2b | 0.57 | 2.74 | 0.48 | 2.32 | 1.21 | 0.998 |
| P3a | 3.21 | 3.15 | 2.38 | 2.34 | 6.98 | 0.995 |
| P3b | 7.17 | 7.16 | 4.77 | 4.76 | 18.48 | 0.989 |
| Average | | 4.33 | | 3.23 | | |

### 3.2 Comparison with Symmetric Gaussian Baseline

To evaluate the added value of modeling asymmetry, the decomposition was repeated using a symmetric Gaussian model as a baseline. In this baseline, the skewness parameter was fixed $\alpha = 0$ for all peaks (yielding standard Gaussian shapes), while keeping the same peak detection and fitting procedures. The symmetric model was applied to the four patterns where it could converge (P1a, P1b, P2a, P2b). For the two most complex profiles (P3a, P3b), the symmetric fit did not converge. This is likely due to their highly overlapping and skewed peaks, which could not be accurately approximated by symmetric shapes. Consequently, these two profiles are excluded from this baseline comparison. The results of the symmetric Gaussian fits for P1a, P1b, P2a, and P2b are summarized in Tables 4 and 5, and key differences are highlighted in Figure 2.

Table 4 lists the fitted parameters for the symmetric case: one can observe that without skewness, the model often compensates by adjusting peak widths. For example, in P2a the symmetric model uses some very broad peaks ($\sigma$ up to 6.65 h) to try to cover the long tail of usage that a skewed peak would handle via asymmetry. In P1a and P1b, some symmetric peak widths became smaller or larger than their SGD counterparts, and baseline $C_{base}$ values shifted slightly.

*Table 4. Symmetric Gaussian Model Parameters*

*Parameters from symmetric Gaussian fits ($\alpha = 0$). Units are consistent with Table 1: $C_{base}$, amplitude in gallons/hour or m³/hour, $\sigma$ in hours, and $\alpha$ is unitless (and fixed at zero here).*

| Pattern ID | N peaks | $C_{base}$ | Amplitude Range | Sigma Range |
|---|---|---|---|---|
| P1a | 5 | 8.23 | (1.08, 13.89) | (0.34, 2.71) |
| P1b | 3 | 6.8 | (2.14, 7.61) | (0.75, 3.21) |
| P2a | 5 | 0 | (0.0, 76.94) | (0.62, 6.65) |
| P2b | 6 | 0 | (0.0, 33.22) | (0.99, 3.62) |

The fit quality of the symmetric models was noticeably worse. As shown in Table 5, errors increased across the board. For P2a (the most asymmetric pattern), the RMSE jumped from 3.54 (with SGD) to 10.43 with the symmetric model – roughly triple the error – and the $R^2$ plummeted from 0.988 to 0.897. P1b, which is a relatively simple pattern, saw $R^2$ drop to 0.867 with symmetric peaks, indicating that even one or two misaligned peaks can degrade the fit significantly. On average, across these four cases, RMSE with the symmetric model was more than double that of the skewed model (see Table 5). The maximum hourly error also increased, by as much as 16–20 m³/h in the worst case (P2a). These results quantitatively demonstrate the value of allowing skewness: many real demand peaks are not perfectly symmetric, and forcing symmetry causes the model to either misalign peaks or use additional peaks to compensate, leading to poorer overall fit.

*Table 5. Performance Metrics – Skewed Gaussian Model, Model performance metrics for the skewed Gaussian fit for each pattern. RMSE, MAE, and maximum error are in the same units as the pattern (gallons/hour or m³/hour); R² is unitless.*

| Pattern ID | RMSE | RMSE as % of mean | MAE | MAE as % of mean | Max Error | $R^2$ |
|---|---|---|---|---|---|---|
| P1a | 0.6 | 4.5 | 0.42 | 3.2 | 1.42 | 0.981 |
| P1b | 0.86 | 9.2 | 0.72 | 7.7 | 1.64 | 0.867 |
| P2a | 10.43 | 21.8 | 8.36 | 17.5 | 23.19 | 0.897 |
| P2b | 1.07 | 5.2 | 0.82 | 4.0 | 2.55 | 0.991 |
| Average | | 10.2 | | 8.1 | | |

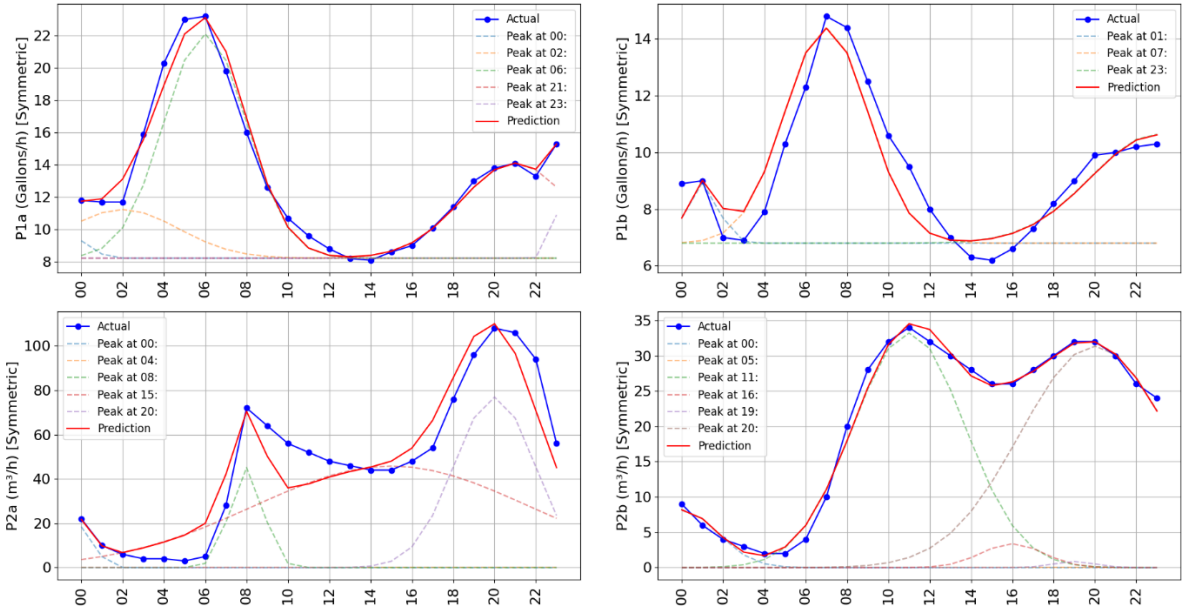

*Figure 2. Observed vs. modeled demand using symmetric Gaussian components (baseline comparison). Plots correspond to the same patterns as in Figure 1 (excluding P3a, P3b). Blue lines are actual demand and red lines are the reconstructed demand using only symmetric Gaussians (α = 0). Green markers indicate detected peak times. By comparing with Figure 1, one can see that the symmetric model struggles in certain cases – for example, in P2a and P2b, the red line deviates from the blue around secondary peaks, and the model cannot capture the extended tail of the morning peak in P2a. This highlights how asymmetric peaks (captured by SGD) yield a better match.*

Overall, Modeling skewness improves reconstruction accuracy and peak representation significantly. The SGD fits showed a wide range of skewness parameters (Table 2), confirming varied shapes in real demand peaks. These improvements were achieved without increasing model complexity, using the same or fewer peaks and only one additional parameter per peak.

# 4  Discussion

This study applies a Skewed Gaussian Distribution (SGD) framework to hourly water demand, enabling decomposition into meaningful components that clarify consumption behavior and system dynamics.

## 4.1  Interpretation of Demand Profiles

The SGD method successfully characterized daily water use across varying complexity levels. Simpler patterns (P1a, P1b) matched typical residential demand cycles with 3–5 peaks representing morning, midday, and evening routines. These align with established findings (Blokker et al. 2010; Di Mauro et al. 2020) and were modeled with high accuracy ($R^2 > 0.98$) using a small number of Gaussian components, supporting earlier conclusions that few components explain most household demand variation. More complex profiles (P3a, P3b) required up to nine components. Though only two or three main peaks were visually evident, the model revealed overlapping sub-peaks likely representing diverse concurrent activities (e.g. industrial, residential, irrigation). This supports the interpretation of aggregate demand as layered (Cominola et al. 2023). COVID-19 profiles (P2a, P2b) highlighted structural shifts: P2a exhibited a sharp morning spike, captured by a high-amplitude, negatively skewed Gaussian, while P2b featured multiple moderate peaks, possibly reflecting staggered routines. These shifts echo observations of altered peak structures during lockdown (Abu-Bakar et al. 2021; Gato et al. 2007), demonstrating the method's sensitivity to behavioral changes. In all three pairs, disruption events produced measurable shifts in SGD parameters: P1b showed a reduction in peak count and intensity post-mandate; P2a revealed sharper asymmetry and higher amplitudes during lockdown; and P3b displayed greater skewness variability and amplitude range, reflecting increased usage heterogeneity.

## 4.2 Operational Interpretability

Each SGD parameter has physical meaning: amplitude ($A$) indicates intensity, location ($\alpha$) pinpoints peak timing, width ($\sigma$) reflects duration, and skewness ($\alpha$) quantifies asymmetry. For instance, a high positive α (e.g. 4.7 in P3b) suggests a rapid onset and gradual decline, commonly observed in residential water use (Beal and Stewart 2014; Cole and Stewart 2013). Conversely, a negative α indicates a gradual buildup and rapid drop-off, often seen during evening-peak hours (Bergel et al. 2017). These parameters expand on previous work (Herrera et al. 2010; McKenna et al. 2014; Pesantez et al. 2020) by quantifying event symmetry. Changes in $\alpha$ or $\mu$ can indicate behavioral or operational changes such as altered routines or appliance use. Baseline flow ($C_{base}$), ranging from near 0 to 24.7 m³/h in the case studies, reflects minimum night demand. This measure serves as a proxy for background leakage or continuous consumption, aligning with night-flow analyses (Behzadian et al. 2014; Romano and Kapelan 2014). Beyond interpretation, isolated peaks allow targeted anomaly detection (McKenna et al. 2014). For example, an unusual spike in $A$ or shift in $\mu$ could signify an event like firefighting or industrial use. Additionally, rising $C_{base}$ or abnormal peak skewness may signal leaks, consistent with strategies emphasizing high-resolution monitoring (Blokker et al. 2010).

## 4.3 Model Comparison and Asymmetry

SGD outperformed symmetric models, especially for sharp or complex peaks (e.g. P2a's morning spike, α ≈ –2). In several cases (P3a, P3b), the symmetric model failed to converge, reinforcing findings from Niknam et al. (2022) and Moretti et al. (2022) that symmetry constraints limit real-world applicability.

By introducing skewness per peak, SGD enabled precise peak alignment, reducing timing errors and improving fit quality. Asymmetry was not rare; skewness values ranged from –3.1 to +4.7. The most asymmetric peaks were also where symmetric models performed worst, confirming the practical necessity of this flexibility.

This study complements prior efforts such as McKenna et al. (2014), who applied Gaussian Mixture Models (GMMs) for demand classification using standard deviation parameters; while their approach focused on clustering typical usage profiles, our method aims to decompose and interpret individual daily patterns in detail, offering peak-level behavioral insights with physical interpretability.

### 4.4 Methodological Scope and Limitations

The framework is diagnostic rather than predictive. Each 24-hour profile was modeled independently using the full dataset, consistent with diagnostic modeling practices(Ghalehkhondabi et al. 2017). Metrics reflect fit quality, not generalization. Only six profiles were analyzed. Though diverse, broader validation would increase confidence in generalizability. Peak detection may be subjective in noisy data; adjustments such as smoothing or threshold tuning can improve reliability. Nonetheless, all tested patterns were handled without failure.

### 4.5 Synthetic Pattern Generation

A key advantage of the SGD model is its capacity to construct synthetic demand profiles using user-defined parameters. This allows scenario simulation in the absence of empirical data. A synthetic weekly pattern was developed to represent urban residential demand in Israel, varying peak timings and amplitudes to reflect weekday and weekend behaviors. Each day featured three peaks - morning, midday, and evening - with variations to simulate cultural routines (e.g. late Friday and Saturday usage). The resulting weekly pattern is shown in Figure 3, which visualizes both the total daily demand and its component Gaussian peaks, highlighting the structural consistency across weekdays and the shifted usage on weekends. Table 6 outlines the parameters used:

*Table 6. Parameters of Synthetic Weekly Water Demand Pattern Generated Using the SGD Model*

| Day | Peaks hours | $C_{base}$ | Amplitudes | Sigmas | Alphas |
|---|---|---|---|---|---|
| Sunday | (7, 12, 19) | 2 | (15, 12, 15) | (3, 4, 2) | (0.1, -0.3, 0.5) |
| Monday | (7, 12, 19) | 2 | (15, 12, 15) | (3, 4, 2) | (0.1, -0.3, 0.5) |
| Tuesday | (7, 12, 19) | 2 | (15, 12, 15) | (3, 4, 2) | (0.1, -0.3, 0.5) |
| Wednesday | (7, 12, 19) | 2 | (15, 12, 15) | (3, 4, 2) | (0.1, -0.3, 0.5) |
| Thursday | (7, 12, 19) | 2 | (15, 12, 15) | (3, 4, 2) | (0.1, -0.3, 0.5) |
| Friday | (8, 15, 20) | 2 | (15, 12, 20) | (3, 3, 2) | (0.5, -0.3, 0.7) |
| Saturday | (9, 17, 21) | 2 | (8, 16, 10) | (3, 5, 3) | (0.5, -0.2, 0.5) |

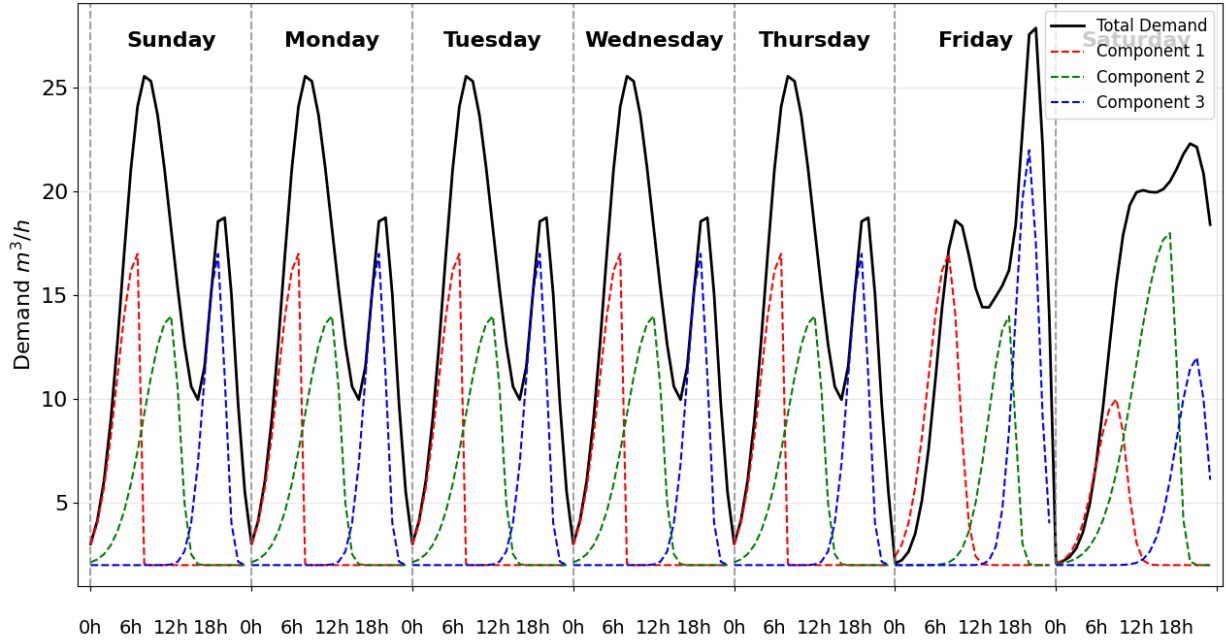

*Figure 3 illustrates the synthetic weekly pattern. The solid line shows total demand; dashed lines represent individual Gaussian components. Sunday through Thursday follows a regular structure, while Friday and Saturday reflect shifted and varied usage patterns.*

This parametric approach enables deliberate scenario design. Adjusting *A*, *μ* or *α* allows simulation of policy impacts (e.g. conservation), infrastructure changes, or routine shifts. Unlike stochastic methods (e.g. SIMDEUM; Blokker et al., 2010), SGD offers full interpretability. Each change maps directly to real-world implications; for example, increasing α leads to a sharper rise and more gradual decline, typical of irrigation or evening leisure usage. All tools and code for real and synthetic analyses are available via the public repository (Elkayam 2025).

## 5    Conclusion

The Skewed Gaussian Decomposition (SGD) method offers a clear and interpretable approach to analyzing daily urban water demand by modeling usage as a combination of baseline flow and skewed peaks. It captures key features - timing, magnitude, duration, and asymmetry - leading to significantly improved accuracy over symmetric models (R² > 0.98). Parameters such as skewness and baseline flow provide valuable operational insights, supporting applications like leak detection, and anomaly identification. The method also enables synthetic demand generation, allowing planners to simulate realistic scenarios for stress testing and future planning. While limited to single-day profiles and not yet predictive, it forms a strong foundation for both retrospective analysis and future integration into forecasting tools.